\pdfoutput=1
\documentclass{article}
\usepackage{spconf,amsmath,graphicx}
\usepackage{enumitem}
\usepackage{adjustbox}
\usepackage{threeparttable}

\usepackage{xcolor,colortbl}

\definecolor{Gray}{gray}{0.85}
\definecolor{LightCyan}{rgb}{0.88,1,1}

\newcolumntype{a}{>{\columncolor{Gray}}c}
\newcolumntype{b}{>{\columncolor{white}}c}

\title{Choose Settings Carefully: Comparing Action Unit detection at Different Settings Using a Large-Scale Dataset}
\name{{Mina Bishay, Ahmed Ghoneim, Mohamed Ashraf and Mohammad Mavadati}}
\address{Affectiva Inc.}
\begin{document}
%
\maketitle
\begin{abstract}



In this paper, we investigate the impact of some of the commonly used settings for (a) preprocessing face images, and (b) classification and training, on Action Unit (AU) detection performance and complexity. We use in our investigation a large-scale dataset, consisting of $\sim$55K videos collected in the wild for participants watching commercial ads. The preprocessing settings include scaling the face to a fixed resolution, changing the color information (RGB to gray-scale), aligning the face, and cropping AU regions, while the classification and training settings include the kind of classifier (multi-label vs. binary) and the amount of data used for training models. To the best of our knowledge, no work had investigated the effect of those settings on AU detection. In our analysis we use CNNs as our baseline classification model. 

\end{abstract}
%


\section{Introduction}
\label{sec:intro}
Automatic Facial Expression Recognition (FER), and in particular Action Unit (AU) detection, is a critical step in analyzing humans’ emotions and non-verbal communication. FER has gained popularity in a wide range of applications, such as ad testing, and driver state monitoring, and social robotics. AU detection has received considerable attention from the Computer Vision community in the last two decades, where different methods have been proposed in the literature \cite{martinez2017automatic, li2020deep}. In these methods, a variety of settings have been used for (a) preprocessing the face images and (b) classification and training. However, it has not been fully investigated how these settings affect the AU detection performance, and the computational complexity.

\begin{figure*} [t!]
  \centering{\includegraphics[width=0.95 \linewidth]{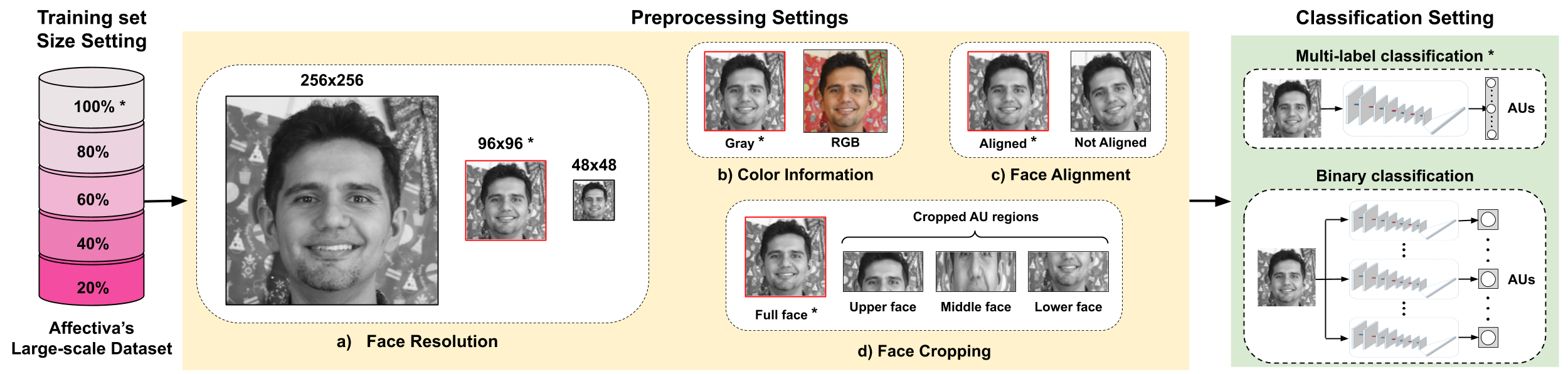}}
  \caption{The different training set size, preprocessing, classification settings compared in this paper ($*$ indicates that this setting belongs to the baseline settings).}
  \label{fig:table_5}
  \label{fig2}
\end{figure*}

In this paper we aim to investigate the effect of some commonly used settings on AU detection performance and complexity. In our investigation we use a large-scale dataset (provided by Affectiva), that consists of a well-diverse set of videos/participants (around 55K videos). This is in contrast to some of the publicly-available datasets, with relatively limited data, recording conditions, and/or variations in demographics. This dataset was captured in the wild and with totally naturalistic facial expressions. Participants’ videos were collected worldwide (from 90+ countries), and annotated for different AUs by trained Facial Action Coding System (FACS \cite{EkmanBook97}) coders. In all our experiments we use CNNs for feature extraction and classification, as CNNs have shown to learn better features than hand-crafted ones \cite{krizhevsky2012imagenet}. The settings we aim to study include preprocessing, classification and training settings (settings are shown in Fig. \ref{fig2}).

\textbf{Preprocessing settings.} The quality and appearance of input face images to the CNNs can vary in different ways. Specifically, the face resolution varies considerably across the literature, typically ranging from small resolutions such as 48$\times$48 \cite{ghosh2015multi, bishay2017fusing, kahou2016emonets} to large ones such as 256$\times$256 \cite{tHoser2016deep, hayale2019facial, vemulapalli2019compact}). The face resolution affects the computational complexity and potentially the model performance. Subsequently, choosing a proper resolution is an important step in designing the AU detection systems. In addition, some researchers \cite{ghosh2015multi, bishay2017fusing, jaiswal2016deep} converted the color (RGB) images to grayscale, while others \cite{zhao2016deep, bishay2019schinet} employed the RGB images. Also, many works localized and aligned the face images before feature extraction \cite{zhao2016deep, bishay2017fusing, hayale2019facial}, while some studies used the images without alignment \cite{bishay2019schinet, jang2017smilenet}. Lastly, some scholars used the full face images as input to the CNNs \cite{ghosh2015multi, bishay2017fusing, hayale2019facial}, while others cropped AU regions \cite{jaiswal2016deep, onal2019d}. The effect of these preprocessing settings needs to be clarified.

\textbf{Classification and training settings.} There are two commonly used approaches for AU classification: \emph{binary classification} where a classifier is trained for each AU \cite{baltruvsaitis2015cross, valstar2015fera}, and \emph{multi-label classification} where a single classifier is trained for predicting all AUs simultaneously \cite{ghosh2015multi, bishay2017fusing}. In binary classification, the model learns AU-specific features, while in multi-label classification, the model learns shared features across the different AUs. Another setting that varies across the literature is the size of the dataset used for training the CNN architecture. Considering all of these variations, there is no existing work which compares and baselines the impact of these preprocessing, classification and training settings on AU detection performance and complexity. The objective of this paper is to clarify the effect of each of these settings, so other researchers can properly consider deploying them as a part of their next modeling.

This paper is organized as follows: Section 2 clarifies the dataset and the baseline settings used in all our experiments. Section 3 compares the different preprocessing, classification, and training settings. Finally, we draw our conclusions in Section 4.



\section{Baseline Settings}

\subsection{Dataset}

In the past decade, several datasets have been collected and annotated for spontaneous behavior, e.g. DISFA \cite{mavadati2013disfa}, DISFA+ \cite{mavadati2016extended}, UNBC \cite{lucey2011painful}, and BP4D \cite{zhang2014bp4d}. However, these datasets have a relatively limited number of subjects, recording conditions, and/or variations in demographics. For our analysis, we collected and labeled a large-scale face video dataset, that was captured in realistic conditions and with totally spontaneous facial expressions. Our dataset contains $\sim$55K face videos with diverse demographics, divided into 40K videos for training, 5K videos for validation, and 8K videos for testing. 


Using a web-based framework \cite{mcduff2013affectiva, mcduff2018fed+}, thousands of videos were collected for individuals watching commercial ads. The participants were recruited from around the world (from 90+ countries). The presence of AUs was manually annotated by trained FACS coders. In addition to the AU annotation, videos were also labeled in terms of gender (55\% Female, 37\% Male, 8\% uncertain) and ethnicity (37\% Caucasian, 24\% East Asian, 14\% South Asian, 13\% Latin, 9\% African, 3\% uncertain). A part of this dataset was made available to the research community through AM-FED \cite{mcduff2013affectiva} and AM-FED+ \cite{mcduff2018fed+}. 

\begin{table*}[t]
\caption{The accuracy and computational complexity across the different AU detection settings.}
\label{table1}
\begin{adjustbox}{width=0.9 \textwidth,center}
\begin{threeparttable}
    \begin{tabular}{|c||a||c|c||c||c||c||c||c|c|c|c|c|}
 \hline
 Exp. & Baseline & 48$\times$48 & 256$\times$256 & RGB & Not & Cropped AU & Binary & 20\% of & 40\% of & 60\% of & 80\% of \\
   & exp. & & & & aligned &  regions  & classification &  videos & videos & videos & videos \\
 
 \hline
  \hline
 AU1 &	75.5 & 69.6 &	76.8 &	76.1 &	70.1 &	78.6 & 76.0 &	52.4 & 60.6 & 60 & 70.1 \\
 \hline
 AU2 &	67.1 &	65.6 &	68.2 & 67.8 &	67.2 &	68.6 & 69.2 &	60.8 & 65.5 & 66.5 & 67.8 \\
 \hline
 AU4 &	68.5 &	67.1 &	72.6 &	69.3 &	68.4 &	75.5 & 73.8 &	64.0 & 66.7 & 68.1 & 69.1 \\
 \hline
 AU6 &	83.5 &	82.5 &	82.7 &	84.7 &	81.9 &	82.1 & 82.8 &	78.6 & 80.1 & 81.1 & 83.6 \\
 \hline
 AU7 &	69.3 &	67.9 &	69.4 &	70.0 &	68.9 &	72.8 & 70.0 &	60 & 63.2 & 65.3 & 69.7 \\
 \hline
 AU9 &	77.2 &	73.5 &	77.2 &	79.0 &	77.4 &	78.0 & 75.4 &	68.2 & 72 & 74.8 & 76.6 \\
 \hline
  AU12 & 84.8 &	83.8 &	85.2 &	85.2 &	84.4 &	84.0 &	85.3 &	80.2 & 83.5 & 84.3 & 84.3 \\
 \hline
 AU15 &	78.6 &	78.3 &	80.0 &	80.0 &	79.0 &	78.9 &	80.6 &	69.4 & 76.1 & 77.4 & 78.4 \\
 \hline
 AU17 &	76.9 &	77.3 &	77.7 &	77.4 &	76.7 &	77.5 &	74.0 &	70.5 & 73.7 & 76.1 & 76.1 \\
 \hline
 AU24 &	77.6 &	76.3 &	77.4 &	77.4 &	76.7 &	77.7 &	75.2 &	67.6 & 73.1 & 76.1 & 77.0 \\
 \hline
 AU25 &	72.7 &	71.4 &	73.9 &	75.5 &	72.3 &	73.0 &	78.0 &	63.6 & 70.4 & 71.3 & 72.0 \\
 \hline
 AU28 &	83.8 &	83.2 &	84.8 &	84.8 &	83.2 &	84.8 &	85.0 &	76.1 & 81.7 & 83.6 & 83.5 \\
 \hline
  \hline
 Avg &	76.3 &	74.7 &	77.2 &	77.3 &	75.5 &	77.6 &	77.1 &	67.1 & 72.2 & 74.8 & 75.9 \\
 \hline
 \hline
 FLOPs & 81.6M$^{+}$ & 21.3M$^{+}$ & 512.3M$^{+}$ & 92.3M$^{+}$ & 81.6M & 124.2M$^{+}$ & 979.2M$^{+}$ & 81.6M$^{+}$ & 81.6M$^{+}$ & 81.6M$^{+}$ & 81.6M$^{+}$ \\
  & & & & & & (41.4M/region)  &  (81.6M/AU) &  &  &  &  \\
 
 \hline
    \end{tabular}
\begin{tablenotes}
\item $^{+}$ indicates that there is an additional complexity coming from the landmark detection architecture, this complexity varies according to the used architecture. Landmarks are used mainly for aligning and cropping the faces.
\end{tablenotes}
\end{threeparttable}
\end{adjustbox}
\end{table*}



\subsection{Architecture}

\textbf{Baseline preprocessing} consists of 3 main steps. We first extract the participant’s full face by using a face detector trained in the wild, and then the detected face is converted to grayscale. Second, we detect 4 facial landmarks on the face; outer eye corners, nose tip and chin. These landmarks are used for eliminating the head roll rotation in the face image, by aligning the eyes horizontally. Finally, the aligned grayscale image is scaled to a fixed resolution of size 96$\times$96, and passed as an input to a CNN. Note that we use the full face as input (no AU-based cropping is applied).



\textbf{Baseline model.} We use a CNN consisting of 4 convolutional and 2 fully-connected layers in our analysis. Each convolutional layer is followed by a max-pooling layer with a filter of size 2$\times$2. The 4 convolutional layers have 32, 32, 64, and 64 filters respectively, and all of the filters have a kernel size of 3$\times$3. We treat the AU detection problem as a multi-label problem where a single CNN is trained for detecting 12 AUs simultaneously (list of AUs is given in Table \ref{table1}). The first fully-connected layer has 256 neurons, while the second has 12 sigmoid units representing the predictions of the 12 AUs.



\subsection{Experimental Settings}

\textbf{Training settings.} In our naturalistic dataset, the frequency of most of the AUs is highly skewed or imbalanced (i.e. having a high ratio of negative to positive frames). In order to balance the positive and negative examples, we use a \textit{balanced sampler} in our analysis. Specifically, for each training batch, we sample 8 images for each AU (i.e. 4 positive and 4 negative images), which results in 96 images for the 12 AUs. The images are sampled randomly from the different training set videos. The total loss is calculated as the average of the independent AU losses (similar to \cite{bishay2017fusing}). It is worth noting that the balanced sampler discards the correlations incorporated between the different AUs, while it retains the AUs equally represented and balanced in the training batches. 






We initialize the CNN randomly from the same seed in all of our experiments. We augment the training batches by random flipping, shearing, scaling, etc. We train the CNN for 600 epochs, using the Adagrad optimizer with an initial learning rate set to 0.005. For each epoch, we sample 300 batches for training and 300 batches for validation. For testing, the best validation model is tested on 3000 batches sampled from the testing set. We use binary cross-entropy for calculating the loss, and accuracy for evaluating performance. The reason we used accuracy for evaluation is that we have an equal number of positive and negative examples in the validation and testing batches. 





\section{Comparing Different Settings}

In this section we will compare the effect of 6 commonly-used settings for preprocessing inputs, building and training CNNs on AU detection performance and complexity, using Affectiva's large-scale dataset. The 6 settings are  shown in Fig. \ref{fig2}. Note that in each of the following experiments we use the same baseline settings, and only change the tested setting.



\subsection{Face Resolution (48$\times$48 -- 256$\times$256)}
\label{res}

The face resolution used as input to the CNNs varies considerably across the literature. Some works used relatively small resolutions such as 48$\times$48 \cite{ghosh2015multi, bishay2017fusing, kahou2016emonets}, while others used large ones such as 256$\times$256 \cite{tHoser2016deep, hayale2019facial, vemulapalli2019compact}). The computational complexity of the CNN increases at higher face resolutions. However, it has not been clearly reported how much the face resolution affects the AU detection performance. In this section, we compare the impact of 3 different resolutions -- 48$\times$48, 96$\times$96, and 256$\times$256. For the 48$\times$48 and 256$\times$256 resolutions, we modify the CNN so as to have the same number of parameters as the 96$\times$96 resolution, specifically by removing or increasing the filter size of the last pooling layer (filter is removed for 48$\times$48 and increased to the size of 4$\times$4 for 256$\times$256). In addition, we remove the padding of the last two convolutional layers for the 256$\times$256 resolution, so as to map all resolutions to the same dimensionality — this will help us compare different resolutions on similar conditions.





Table 1 shows the performance and computational complexity for the 48$\times$48, 96$\times$96 (baseline experiment), and 256$\times$256 experiments. Results show that increasing the resolution increases both the average accuracy and computation complexity of the models. More specifically, moving from the low 48$\times$48 to the 96$\times$96 resolution increases the accuracy by 1.6\% and the number of FLOPs by $\sim$4 times. Also, moving from the 96$\times$96 to the large 256$\times$256 resolution increases the performance by $\sim$1\% and the number of FLOPs by $\sim$6 times. Therefore, we advise researchers to carefully select the face resolution, keeping in mind the trade-offs in performance and processing time.






\subsection{Color Information (Gray-scale vs. RGB)}

In the literature, some researchers relied solely on single input channel (i.e. gray-scale face images) for AU detection \cite{ghosh2015multi, jaiswal2016deep, bishay2017fusing}, while others used color (RGB) images \cite{zhao2016deep, bishay2019schinet}. Using RGB images increases the computational complexity of the CNN model — in our case the number of FLOPs increased by $\sim$13\%. It has not been fully investigated how much the additional color information can increase the model performance. In this section, we compare the impact of using the gray-scale and RGB images in AU detection. 



Table \ref{table1} shows the accuracy and computational complexity for the gray-scale (baseline experiment) and RGB input type experiments. Using RGB images shows on average 1\% improvement in accuracy while adding $\sim$13\% extra computations. The improvement can be seen across most of the AUs (e.g. AU6, AU25). Hence, we encourage the community to use RGB images in AU detection, especially with deeper CNNs, as the difference in computations between the gray-scale and RGB experiments gets lower with deeper CNNs.

\subsection{Face Alignment (Aligned vs. Not-Aligned)}

Many papers across the literature used alignment as a pre-processing step in AU detection \cite{zhao2016deep, bishay2017fusing, hayale2019facial}, in which the facial landmarks of the eye corners are used to align the face to have a zero roll angle. Aligning images increases the computational cost of the AU detection pipeline, and this increase depends on the complexity of the landmark detection architecture. In \cite{jang2017smilenet}, Jang {\em et al.} proposed not to use alignment in smile detection. In this section, we aim to extend that experiment to a wider range of facial expressions, and discover how the face alignment may affect the AU detection performance.



Table \ref{table1} shows the accuracy for experiments based on aligned (baseline experiment) and not-aligned faces. Results show that on average using alignment improves the accuracy by 0.8\%. AUs characterized with subtle change in appearance like AU1 and AU6 are the ones that got notably improved by alignment, while the rest of the AUs have very close performance. Subsequently, alignment is only recommended when detecting those specific AUs, otherwise alignment can be skipped as this might save computations, by not using the landmark detection architecture.

\subsection{Face Cropping (Full Face vs. AU Regions)}

Several studies used the full face as input to the CNN \cite{ghosh2015multi, bishay2017fusing, hayale2019facial}, while others used the region of the face relevant to each AU for AU detection \cite{jaiswal2016deep, onal2019d}, as this allows the model to learn more localized features for each AU. The region-based analysis typically uses facial landmarks to define the region of interest for each AU, and this definition can differ from one study to another. In this section, we compare how using the full face and the cropped AU regions affects the AU detection performance. For region-based analysis, we divide the full face into 3 regions; lower, middle, and upper face. Then, the cropped regions are scaled to a fixed resolution of 48$\times$96. Fig. \ref{fig2} shows an example for 3 cropped regions. We assign AU1, AU2, AU4, AU7 and AU9 to the upper face, AU6 to the middle face, and AU12, AU15, AU17, AU24, AU25, and AU28 to the lower face region.


For region-based AU detection, we train 3 CNNs, one for each cropped region. In order to have a comparable number of parameters as the baseline/full-face setting, we reduce the filter size of the last pooling layer. Table \ref{table1} reports the accuracy and complexity for the full face (baseline experiment) and the cropped AU regions experiments. Results show that the region-based analysis improves the accuracy on average by 1.3\%. Improvement is seen across most of the AUs. However, the region-based analysis increases the computational complexity as 3 CNNs have been used, and this complexity might increase more if the face is cropped to more regions.

\subsection{Kind of Classifier (Multi-label vs. Binary)}
\label{BvM}

In the literature, many works dealt with AU detection as independent binary classification problems \cite{baltruvsaitis2015cross, valstar2015fera}, where a different classifier is trained for each AU, and subsequently, each classifier is learning AU-specific features. The computational complexity of such an approach increases linearly as the number of detected AUs increases. Other works dealt with AU detection as a multi-label classification problem \cite{ghosh2015multi, bishay2017fusing}, where a single classifier is trained jointly for the detection of all AUs, and subsequently the architecture is learning shared features across the different AUs. In this section, we compare how using the multi-label and binary settings affect the AU detection performance. 


Table \ref{table1} shows the accuracy and computational complexity for the multi-label (baseline experiment) and binary settings. Results show that using the binary setting improves the overall accuracy by 0.8\%. However, the accuracy varies across the different AUs, where 8 AUs got improved by using the binary setting, while 4 AUs showed better results by using the multi-label setting. Subsequently, researchers are encouraged to use binary classification for AUs with relatively large improvement like AU4, and multi-label classification for other AUs, as the computational cost for the binary setting is high.

\subsection{Training Set Size}

Several datasets have been annotated in terms of FACS and used for AU detection — these datasets have different numbers of subjects and videos/images. In this section, we explore how the size of the training set (i.e. number of labeled videos/subjects) affects the AU detection performance. Specifically, we train a CNN using 20\%, 40\%, 60\%, 80\%, and 100\% (baseline experiment) of our training set videos, and then explore how much the performance changes. Note that the validation and testing sets are kept the same for all the experiments.  Table \ref{table1} shows that the more videos we add to the training set, the better the accuracy. However, each time we add 20\% more videos, accuracy does not improve with the same rate, that is, getting closer to 80\% or 100\% adds a small improvement to the overall accuracy, while moving from 20\% to 40\% or 40\% to 60\%, adds relatively larger improvement to the accuracy.



\section{Conclusion}

In this work, we investigated the impact of some commonly-used settings for preprocessing, building and training CNNs, on AU detection performance and complexity. We use in our analysis a large-scale dataset consisting of $\sim$55K videos from different demographic groups. Results show that using a) large resolution, b) alignment, c) color images, d) region-based analysis, and e) binary classification led to improvement in the overall accuracy (by $\sim$1-2\% for each setting), however each setting has an additional computational cost ($\sim$1.1-12 times the baseline cost). In addition, training AU detection architectures requires a large amount of data. Subsequently, those settings need to be carefully chosen based on the available computational power for each application.




\section{Acknowledgments}

We would like to thank Jay Turcot and Dr. Rana el Kaliouby for their continued support and technical guidance, and Steve Hunnicutt for proof-reading the paper and his valuable constructive feedback.


\bibliographystyle{IEEEbib}
\bibliography{ref}


\end{document}